\documentclass{article}

    \usepackage[preprint, nonatbib]{neurips_2024}

\usepackage[nonatbib]{neurips_2024}
\usepackage[numbers]{natbib}
\usepackage[utf8]{inputenc} 
\usepackage[T1]{fontenc}    
\usepackage[hidelinks,colorlinks=true,linkcolor=blue,citecolor=blue]{hyperref}       
\usepackage{url}            
\usepackage{booktabs}       
\usepackage{amsfonts}       
\usepackage{nicefrac}       
\usepackage{microtype}      
\usepackage{xcolor}         
\usepackage{graphicx}

\usepackage{amsmath, amssymb, amsthm}
\usepackage{bm}
\usepackage{mathtools}

\renewcommand{\vec}[1]{\boldsymbol {#1}}

\newcommand{\rvec}[1]{\boldsymbol{\mathrm{#1}}}

\newcommand{\bb}[1]{\boldsymbol{#1}}
\newcommand{\T}[0]{{^\top}}

\DeclareMathOperator*{\argmin}{arg\,min}

\usepackage{authblk}        
\usepackage{cleveref} 
\usepackage{caption}
\usepackage{subcaption}
\usepackage{booktabs}       
\usepackage[ruled]{algorithm2e}
\usepackage{wrapfig}
\usepackage{sidecap}

\title{Scalable unsupervised alignment of general metric and non-metric structures}

\author[1,2]{\textbf{Sanketh Vedula}}
\author[2,3]{\textbf{Valentino Maiorca}}
\author[2,4]{\textbf{Lorenzo Basile}}
\author[2,$\dagger$]{\\\textbf{Francesco Locatello}}
\author[1,2,$\dagger$]{\textbf{Alex Bronstein}}

\affil[1]{Technion -- Israel Institute of Technology}
\affil[2]{Institute of Science and Technology, Austria}
\affil[3]{Sapienza University of Rome}
\affil[4]{University of Trieste}
\affil[$\dagger$]{Equal advising.}

\begin{document}

\maketitle

\begin{abstract}
\looseness=-1Aligning data from different domains is a fundamental problem in machine learning
 with broad applications across very different areas, most notably aligning experimental readouts in single-cell multiomics.
Mathematically, this problem can be formulated as the minimization of disagreement of pair-wise quantities such as distances and is related to the Gromov-Hausdorff and Gromov-Wasserstein distances. Computationally, it is a quadratic assignment problem (QAP) that is known to be NP-hard. 
Prior works attempted to solve the QAP directly with entropic or low-rank regularization on the permutation, which is computationally tractable only for modestly-sized inputs, and encode only limited inductive bias related to the domains being aligned. 
We consider the alignment of metric structures formulated as a discrete Gromov-Wasserstein problem and instead of solving the QAP directly, we propose to \textit{learn} a related well-scalable linear assignment problem (LAP) whose solution is also a minimizer of the QAP.
We also show a flexible extension of the proposed framework to general non-metric dissimilarities through differentiable ranks. 
We extensively evaluate our approach on synthetic and real datasets from single-cell multiomics and neural latent spaces, achieving state-of-the-art performance while being conceptually and computationally simple. 
\end{abstract}

\section{Introduction}

Unsupervised alignment of data that are related, yet not directly comparable, is a fundamental problem in machine learning. This problem is ubiquitous across a multitude of tasks such as \textit{non-rigid shape correspondence} in computer vision~\citep{bronstein2006generalized, halimi2019unsupervised}, \textit{unlabeled sensing} in signal processing~\citep{unnikrishnan2018unlabeled, emiya2014compressed}, and \textit{latent space communication} in representation learning~\citep{moschella2022relative, maiorca2024latent}. From an application perspective, we are particularly interested in single-cell biology.
In fact, the development of single-cell sequencing technologies has led to the profiling of different molecular aspects within the cell at an unparalleled resolution. Profiling techniques have been developed to assay gene expression~\cite{sc_gene_expression}, chromatin accessibility and 3D conformation~\cite{sc_chromatin_accessability, sc_chromatin_conformation_accessibility}, DNA methylation~\cite{sc_methylation}, and histone modifications~\cite{sc_histone_sequencing}.
The analysis of genome~\cite{sc_genomic_analysis_helps1, sc_genomic_analysis_helps2}, transcriptome~\cite{sc_transcriptome_analysis_helps1, sc_methylation_profiling_helps2}, and DNA methylation~\cite{sc_methylation_profiling_helps1, sc_methylation_profiling_helps2} profiles has led to enhanced understanding of the heterogeneity across cell populations. The development of high-throughput sequencing~\cite{high_throughput_1, high_throughput_2, high_throughput_3}, and spatial transcriptomics~\cite{spatial_transcriptomics} technologies further enabled molecular profiling of cells at a high temporal and spatial resolution.
One of the central problems within {single-cell multiomics} is integrating data from different molecular profiles, which is crucial in understanding joint regulatory mechanisms within the cell. Most single-cell sequencing techniques are invasive; thus, carrying out multiple assays on the same cell is rarely possible.
While experimental co-assaying techniques are an active area of research~\cite{sc_multiomics_technique, sc_multiomics_survey}, they currently lack the high throughput of their single-assay counterparts. 
Computationally integrating data from different experimental modalities is, therefore, an important problem, and is the focus of the current paper.

Using the formalism of Gromov-Hausdorff (GH)~\cite{gromov1999metric} and Gromov-Wasserstein (GW)~\cite{memoli2011gromov} distances, unsupervised alignment can be formulated as the minimization of disagreement in pair-wise distances. 
Given two point clouds, both the GH and GW problems aim to find an \textit{assignment} that is invariant to distance-preserving transformations (isometries) of the point clouds. GH seeks an \textit{exact} point-wise assignment and can be shown to be a quadratic assignment problem (QAP) that is known to be an NP-hard~\cite{burkard1998quadratic} and, thus, computationally intractable. GW relaxes the GH problem to find a \textit{soft assignment} and it is more tractable in practice. The most common approach to solving QAP relaxations like GW is by solving a sequence of \textit{linear assignment problems} (LAPs)~\cite{qap_as_sequence_of_laps_works_well} or 
\textit{entropy-regularized optimal transport} ($\epsilon$-OT)~\cite{cuturi2013sinkhorn} problems. This approach, coupled with the idea of kernel matching and, specifically, simulated annealing of kernel matrices, has been demonstrated very successful in shape analysis, practically rendering non-rigid shape correspondence a solved problem~\cite{vestner2017efficient, melzi2019zoomout}. For more general, less structured and higher-dimensional data, recent works have aimed to accelerate the GW solver by (i) reducing the problem size by applying \textit{recursive clustering}~\cite{blumberg2020mrec} or through the \textit{quantization} of the input dissimilarities~\cite{chowdhury2021quantized}; and (ii) imposing \textit{low-rank constraints} on the pairwise distance matrices and the assignment matrix within the internal $\epsilon$-OT solver~\cite{scetbon2022linear}. 

Specifically on the problem of unsupervised alignment of single-cell multiomic data, GW solvers have already shown promise. Nitzan \emph{et al.}~\cite{nitzan2019gene} showed that they could map spatial coordinates in 2D tissues that were obtained with fluorescence in situ hybridization (FISH) to gene expression data. More recently, Demetci \emph{et al.} \cite{demetci2022scot} demonstrated that GW solvers outperform other unsupervised alignment approaches on real data generated by the SNAREseq assay~\cite{chen2019high}, which links chromatin expression to gene expression. Unfortunately, existing solvers have several limitations, including poor scalability to very large ($N\sim 10^4$) datasets, convergence to local minima, and lack of inductivity in the sense that the solver has to be run anew once new data are obtained. This paper proposes remedies to these shortcomings.

\textbf{Contributions. \ } In this work, we introduce a new \textit{framework} for solving GW-like problems. The core idea of our approach is to \textit{learn the cost} of an OT problem (essentially, a LAP) whose solution is also the minimizer of the GW problem (essentially, a QAP). Instead of \textit{explicitly} learning the cost matrix for the given set of samples, we propose to \textit{implicitly} parametrize the cost as a ground-cost measured on neural network embeddings of the points that are being aligned. In order to learn the the neural networks parametrizing the cost, we render the entropy-regularized OT problem as an implicitly differentiable layer using the methodology proposed in \cite{eisenberger2022unified}, and demand that the soft assignment produced by $\epsilon$-OT minimizes the GW objective. 

\looseness=-1This framework offers unique advantages over the standard approach of solving GW as a sequence of LAPs. Firstly, our method is \textit{inductive}. Since we implicitly parametrize the cost with neural networks, when we encounter new pairs of unaligned samples at inference, we simply need to solve an $\epsilon$-OT problem on the embeddings produced by our trained network. This is in contrast to all the other GW solvers that, to the best of our knowledge, are \textit{transductive} and would need to solve the GW problem anew by augmenting the test points. Secondly, our framework is \textit{scalable} requiring to only solve a point-wise $\epsilon$-OT problem at inference. Compared to GW, $\epsilon$-OT is far simpler, and efficient solvers can be employed to solve this problem at scale~\cite{cuturi2013sinkhorn, genevay2016stochastic}. Thirdly, our framework is gradient descent-based and is, therefore, \textit{more expressive and general}, as it is straightforward to induce additional domain knowledge into the problem or impose additional regularization on the minimizer. Furthermore, it is straightforward to extend our method to the semi-supervised setting where a partial correspondence is known, and to the \textit{fused GW}~\cite{fusedgw} setting where a shared attribute is provided in both domains.

Leveraging the advantages of the proposed framework, we propose several novel extensions.
Firstly, we demonstrate, for the first time, solving \textit{arbitrary non-metric} assignment problems. To this end, we propose a new objective that matches distance ranks instead of the absolute distances themselves and demonstrate that it is more effective in single-cell multiomic alignment. The standard GW solvers rely on the linearization of QAPs, and it is unclear how they can be extended to handle more complex objectives such as those involving ranking.
Secondly, inspired by techniques in geometric matrix completion~\cite{kalofolias2014matrix, boyarski2022spectral}, by interpreting the learned cost as a signal on the product manifold of both domains, we impose a regularization that demands that the cost is smooth on its domain. This is intuitive because similar samples in one domain incur a similar cost with respect to the samples from the other domain, and vice-versa. 
Thirdly, in order to robustify training through $\epsilon$-OT solvers, we propose a simulated-annealing--based approach allowing tuning of the regularization coefficient in the Sinkhorn algorithm during the training process. 

We evaluate our method both in inductive and transductive settings, on synthetic and real data. We demonstrated in an inductive setting our solver generalizes and scales to large sample sizes.
We demonstrate that it outperforms the entropic GW solver on the SNARE-seq data from~\cite{snareseq} and on human bonemarrow scATAC vs. scRNA mapping task proposed in~\citet{moscotdata}.

\section{Background and closely related works in Optimal Transport}\label{background}
 Given two sets of points $\{\mathbf{x}_1, \mathbf{x}_2, \ldots \mathbf{x}_N\} \in \mathcal{X}$ and $\{ \mathbf{y}_1, \ldots \mathbf{y}_N \} \in \mathcal{Y}$, the goal of unpaired alignment is to find a \textit{point-wise correspondence} $\mathbf{P} \in \mathcal{P}^N$ such that each point in $\mathcal{X}$ is mapped to a point in $\mathcal{Y}$, and vice-versa, where $\mathcal{P}^N$ is the space of permutations. The central theme of metric-based alignment approaches (GH and GW) is to compare the sets of points as \textit{metric spaces}. $\mathcal{X}$ and $\mathcal{Y}$ are considered similar if the metrics between corresponding points, as defined by $\mathbf{P}$, are similar as measured in $\mathcal{X}$ and in $\mathcal{Y}$. Denote by $d_\mathcal{X}$ and $d_\mathcal{Y}$ the metrics associated to $\mathcal{X}$ and $\mathcal{Y}$, and by
$\mathbf{D}_\mathcal{X} \in \mathbb{R}^{N \times N}$ and $\mathbf{D}_\mathcal{Y} \in \mathbb{R}^{N \times N}$ the corresponding pairwise distance matrices computed over the points from $\mathcal{X}$ and $\mathcal{Y}$, respectively. Let further $\mu$ and $\nu$ be the associated discrete probability measures on $\mathcal{X}$ and $\mathcal{Y}$, respectively. Depending on what the spaces represent, these can be uniform measures or incorporate discrete volume elements. 

\paragraph{Gromov-Hausdorff distance.} The \textit{distortion} induced by a correspondence $\mathbf{P}$ between $(\mathcal{X}, d_\mathcal{X})$ and $(\mathcal{Y}, d_\mathcal{Y})$ is defined as $
\text{dis}(\mathbf{P}) =  \| \mathbf{D}_\mathcal{X} - \mathbf{P} \mathbf{D}_\mathcal{Y} \mathbf{P}\T \|_\infty.
$
This measures how well the distances between the matched points are preserved.
The Gromov-Hausdorff (GH) distance~\cite{gromov1999metric} is then defined as 
\begin{equation}\label{eq:GH}
    d_{\text{GH}}((\mathcal{X}, d_\mathcal{X}), (\mathcal{Y}, d_\mathcal{Y})) = \min_{\mathbf{P} \in \mathcal{P}^N} \text{dis}(\mathbf{P}).
\end{equation}
The optimization problem in Eq. \ref{eq:GH} results in an integer linear program and is an NP-hard problem~\cite{burkard1998quadratic}. Therefore, it is computationally intractable.

\paragraph{Gromov-Wasserstein distance.} \citet{memoli2011gromov} proposed relaxing the constraint on $\mathbf{P}$ from an exact assignment defined over $\mathcal{P}^N$ to a \textit{probabilistic} (soft) assignment, i.e., to the space of couplings with marginals $\mu$ and $\nu$ denoted by $U(\mu, \nu):= \{ \mathbf{\Pi} \in \mathbb{R}_+^{N \times N} \, | \, \mathbf{\Pi} \Vec{1}_N = \bb{\mu}, \mathbf{\Pi}^\T \Vec{1}_N = \bb{\nu} \}$. Using this relaxation, the squared Gromov-Wasserstein distance between discrete metric spaces is defined as
\begin{equation}
\label{eq:gw}
    d^2_{\text{GW}} = \min_{\mathbf{\Pi} \in U(\mu, \nu)} \sum_{i, j, i', j'} \left(d_\mathcal{X}(\mathbf{x}_i, \mathbf{x}_{i'}) - d_\mathcal{Y}(\mathbf{y}_{i}, \mathbf{y}_{i'})\right)^2 \pi_{ij}\pi_{i'j'} = \min_{\mathbf{\Pi} \in U(\mu, \nu)} \|\mathbf{D}_\mathcal{X} - \mathbf{\Pi} \mathbf{D}_\mathcal{Y} \mathbf{\Pi}\T\|_\mathrm{F}^2.
\end{equation}
To avoid confusion, we reserve the notation $\mathbf{P}$ to the true permutation matrix, while denoting the ``soft" assignment by $\mathbf{\Pi}$.
Notice that the definition of the GW distance results in a quadratic function in $\mathbf{\Pi}$; thus, it is referred to as the \textit{quadratic assignment problem}. Alternative relaxations to the GH problem exist based on semi-definite programming (SDP) ~\cite{gh_sdp}, but due to the poor scalability of SDP problems, they do not apply to the scales discussed in this paper.

\looseness=-1\paragraph{Optimal transport.} Aligning data that lie \textit{within the same space} is a \textit{linear} optimal transport (OT) problem~\cite{peyre2019computational}. Given two sets of points $\{\mathbf{x}_i\}_{i=1}^N$ and $\{ \mathbf{x}'_j \}_{i=1}^N$ in the same space $\mathcal{X}$ with two discrete measures $\mu$ and $\nu$, respectively, the OT problem is defined as the minimization of $\sum_{i, j} \pi_{i,j} c(\mathbf{x}_i, \mathbf{x}'_j)$, such that $\mathbf{\Pi}$ satisfies marginal constraints $U(\mu, \nu)$ and $c$ defines transport cost (often, $c(\bb{x}, \bb{x}') = d_\mathcal{X}(\bb{x}, \bb{x}')$). Note that the objective is \textit{linear} in $\mathbf{\Pi}$, in contrast to GW (Eq. \ref{eq:gw}), where it is quadratic. 

Entropy-regularized OT ($\epsilon$-OT) introduces an entropic regularization term, $\epsilon \langle \mathbf{\Pi}, \log \mathbf{\Pi} \rangle$, that can be very efficiently solved using the Sinkhorn algorithm \cite{cuturi2013sinkhorn} (see Appendix for details). More recently, \citet{eisenberger2022unified} introduced \textit{differentiable Sinkhorn layers} that uses implicit-differentiation~\cite{amos2017optnet} to cast the Sinkhorn algorithm as a differentiable block within larger auto-differentiation pipelines. They calculate the Jacobian of the resulting assignment matrix with respect to both the primal and dual variables of the entropic-regularized OT problem. While $\epsilon$-OT solvers (minimizing a point-wise loss) cannot directly solve the GW problem with its pair-wise loss, it is a crucial building block in the most efficient GW solver existing today, which is described below.
 
\paragraph{Entropic Gromov-Wasserstein.} In a similar spirit to $\epsilon$-OT, \citet{solomon_entropic_QAP} proposed to solve an entropy-regularized version of GW problem (Eq. \ref{eq:gw}).  \citet{peyre2019computational} introduced a mirror-descent-based algorithm that iteratively linearizes the objective in Eq. \ref{eq:gw} and then performs a projection onto $U(\mu, \nu)$ by solving an $\epsilon$-OT problem to obtain an assignment (see Appendix for details). This procedure is repeated for a number of iterations. Since each outer iteration involves solving an OT problem in the projection step, this quickly becomes expensive and intractable even in moderate sample sizes. In our experiments, we observed that entropic GW solvers result in out-of-memory for $N>25000$ even when running on optimized implementation from \texttt{ott-jax}~\cite{cuturi2022optimal} on a high-end GPU, whereas the implementation in \texttt{POT}~\cite{flamary2021pot}, since it is CPU-based, is intractable already for $N>8000$.
\citet{scetbon2022linear} proposed \textit{low-rank GW} that imposes low-rank constraints both on the cost and assignment matrices as an alternative to entropic GW and demonstrated that it could provide speed-up compared to entropic counterpart. We observed that if the data violates the low-rank assumptions, as is generally true for distance matrices and was specifically the case in our real data experiments, the benefits from this approach become void. Explicitly imposing low-rank constraints led to a severe degradation in the quality of the estimated assignment.

\section{Our approach to the GW problem}
\begin{SCfigure}[][t]
    \centering
    \includegraphics[width=0.5\textwidth]{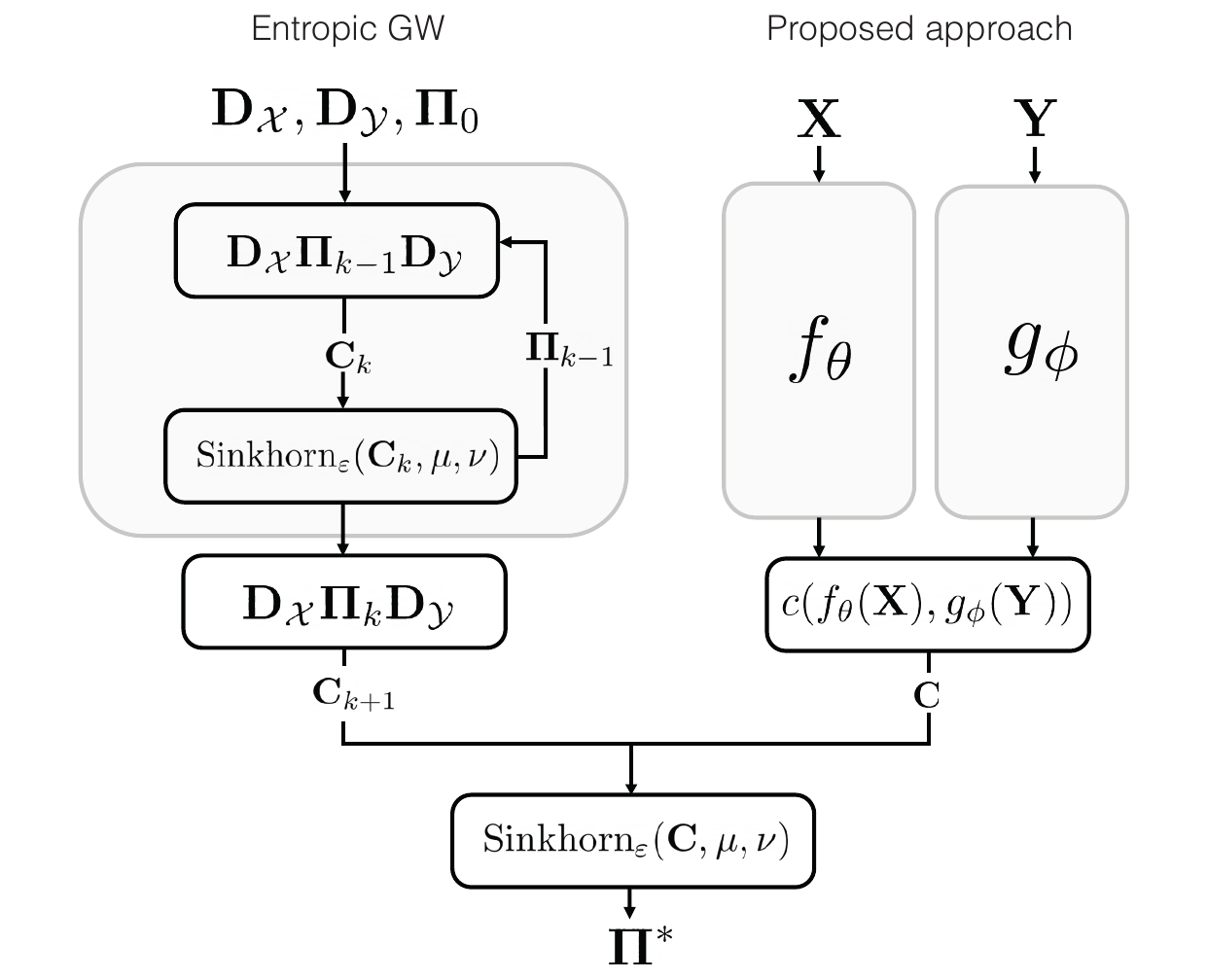}
    \caption{Entropic Gromov-Wasserstein solver (left) solves a sequence of regularized optimal transport ($\epsilon$-OT) problems using the Sinkhorn algorithm. In contrast, the proposed approach learns, via a pair of embeddings, $f_\theta$ and $g_\phi$, the transport cost that directly produces the sought alignment $\vec{\Pi}^\ast$ by solving a single $\epsilon$-OT problem. While the learning of the embeddings still requires multiple calls to the $\epsilon$-OT solver, their cost is amortized at inference time. }
    \label{fig:our_approach_vs_gw}
\end{SCfigure}

\looseness=-1In order to scale GW solvers to large sample sizes, we start with the following question: can we find an entropic OT problem whose solution coincides with that of the entropic GW problem (Eq. \ref{eq:gw})? The rationale is that, given an unpaired set of samples, if we determine an equivalent entropic OT problem, we can employ fast entropic OT solvers to calculate the assignment. One obvious problem that fits this criterion, by construction, is the entropic OT problem that is solved in the last iterate of the entropic GW solver. However, computing this problem would require iterating through the GW solver, and it is thus impractical. By phrasing this question as an optimization problem, we get the following,
\begin{equation}
\begin{aligned}
    \label{eq:ideal_problem_ours}
    \Vec{\Pi}^* =& \argmin_{\mathbf{C}}  \, \left\| \mathbf{D}_\mathcal{X} - \Vec{\Pi}(\mathbf{C}) \mathbf{D}_\mathcal{Y} \Vec{\Pi}^\T(\mathbf{C}) \right\|_{\mathrm{F}}^2 \\
    & \,\, \text{  s.t.  } \Vec{\Pi}(\mathbf{C}) = \argmin_{\Vec{\Pi} \in U(\mu, \nu)} \langle \mathbf{\Pi}, \mathbf{C}\rangle.
\end{aligned}
\end{equation}
It is a bilevel optimization problem: the inner problem is linear OT and it produces an assignment that is optimal with respect to the cost $\mathbf{C}$, and the outer problem demands that the resulting $\mathbf{\Pi}(\mathbf{C})$ is GW-optimal, i.e., it aligns the metrics $\mathbf{D}_\mathcal{X}$ and $\mathbf{D}_\mathcal{Y}$. While seemingly elegant, Equation \eqref{eq:ideal_problem_ours} has two major problems: (i) because $\mathbf{C}$ is unbounded, this objective is very unstable and difficult to optimize; (ii) more practically, Eq. \ref{eq:ideal_problem_ours} results in a \textit{transductive} approach; given a new set of unpaired samples, this problem needs to be solved anew, which is not scalable. 

To mitigate this, instead of optimizing the cost matrix $\mathbf{C}$ (in Eq~\ref{eq:ideal_problem_ours}), we propose to \textit{implicitly}  parametrize it as a pairwise cost measured on the \textit{learned embeddings} of pointwise features $\mathbf{X}$ and $\mathbf{Y}$. This leads us to the following modified objective,
\begin{equation}
\begin{aligned}
    \label{eq:final_objective_ours}
    \Vec{\Pi}^* =& \argmin_{\theta, \phi}  \, \left\| \mathbf{D}_\mathcal{X} - \Vec{\Pi}(\theta, \phi) \mathbf{D}_\mathcal{Y} \Vec{\Pi}^\T(\theta, \phi) \right\|_{\mathrm{F}}^2 \\
    & \,\, \text{  s.t.  } \Vec{\Pi}(\theta, \phi) = \argmin_{\Vec{\Pi} \in U(\mu, \nu)} \langle \mathbf{\Pi}, {c}(f_\theta(\rvec{X}), g_\phi(\rvec{Y}))\rangle,
\end{aligned}
\end{equation}
where $f, g$ are learnable functions, modeled via neural networks, embedding $\mathbf{X}$ and $\mathbf{Y}$, respectively. It is important to emphasize that the cost is realized through the embedding, while the function $c$ is fixed to the simple Euclidean ($c(\bb{z}, \bb{z}') = \| \bb{z} - \bb{z}' \|^2$) or cosine ($c(\bb{z}, \bb{z}') = \bb{z}\T \bb{z}'$) form. We solve the above problem via gradient descent. In order to backpropagate gradients to $f$ and $g$, we first relax the inner problem to be an $\epsilon$-OT problem, and then employ implicit differentiation~\cite{amos2017optnet} to calculate $\frac{\partial{\mathbf{\Pi}}}{\partial{c}}$~\cite{eisenberger2022unified} which is backpropagated to update the weights of $f$ and $g$.

From a geometric perspective, we are embedding the samples from $\mathcal{X}$ and $\mathcal{Y}$ into a common domain $\mathcal{Z}$, where the samples are \textit{OT-aligned} with the same assignment that makes the metric spaces $(\mathcal{X}, d_\mathcal{X})$ and $(\mathcal{Y}, d_\mathcal{Y})$ \textit{GW-aligned}. 
From a \textit{computational} point of view, our framework can be viewed as an \textit{amortized entropic GW solver}. Figure \ref{fig:our_approach_vs_gw} presents the parallels between our solver and the entropic GW solver~\cite{solomon_entropic_QAP}. 
The ground cost of measured on the embeddings $c(f_\theta(\mathbf{X}), g_\phi(\mathbf{Y}))$ can be interpreted as the cost matrix $\mathbf{C}_{k+1} = \mathbf{D}_\mathcal{X} \Vec{\Pi}_k \mathbf{D}_\mathcal{Y}$ (as depicted in the Fig.~\ref{fig:our_approach_vs_gw}) produced by running the entropic GW solver for $k$ iterations. Post training, the neural networks can be viewed to be amortizing the GW iterations, in similar spirit to recent amortized optimization techniques proposed for fast calculation of convex conjugates~\cite{amos2023tutorial, amos2022amortizing}.

From a practical standpoint, this results in an \textit{inductive} GW solver. At inference, when a new set of unpaired samples from $\mathcal{X}$ and $\mathcal{Y}$ are encountered, we simply need to solve an entropic OT problem that is highly scalable. Moreover, since our solver is gradient-descent-based, it allows the flexibility to induce domain knowledge, additional regularization, and inductive biases on the assignment, on the cost, and in the neural networks $f, g$, respectively. We will discuss a few such examples in the sequel.
Finally, while our approach may resemble the \textit{inverse OT} (iOT) problem~\cite{inv_ot1, inv_ot2} in the sense that it involves the learning of the transport cost, it greatly differs in the minimized objective. While iOT targets finding a cost realizing a given assignment (hence, requiring coupled data), our learning problem does not assume a known target permutation; instead, it tries to find one minimizing the pairwise distance disagreement on unaligned data. From this perspective, the proposed approach can be seen as a variational analog of the iOT problem. 

\begin{figure}
    \centering
    \includegraphics[width=0.4\textwidth]{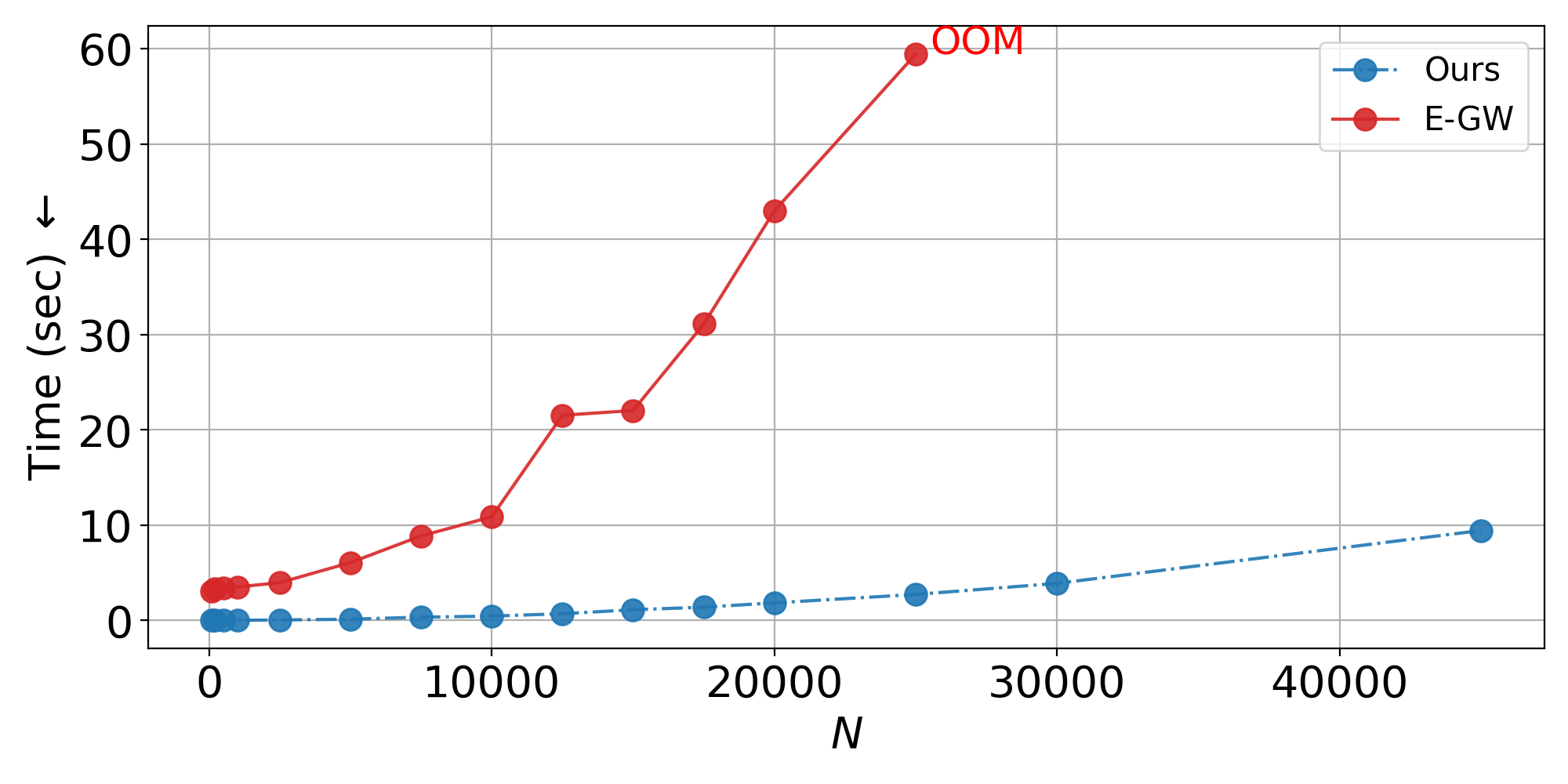}
    \includegraphics[width=0.4\textwidth]{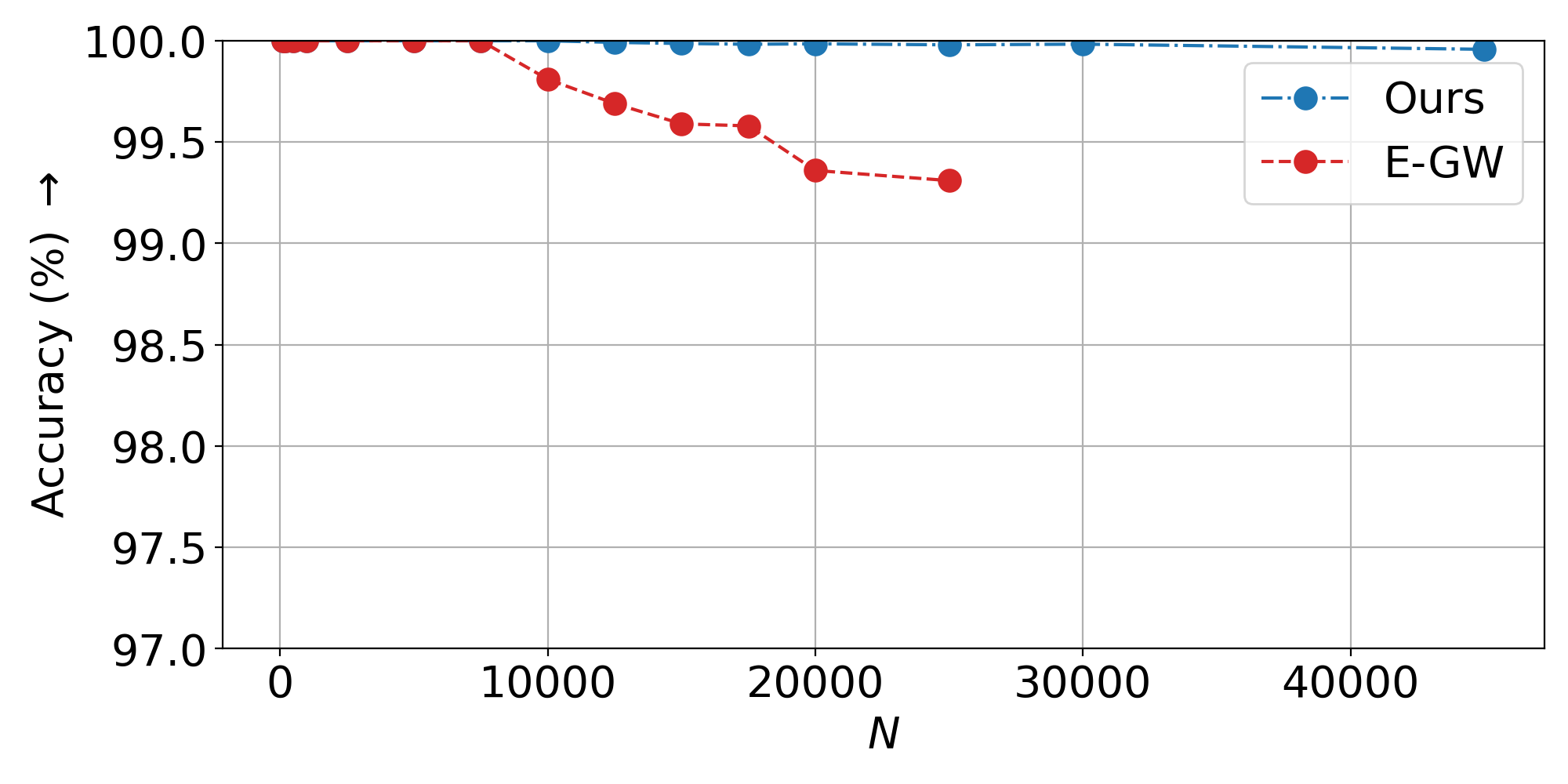}\\
    \includegraphics[width=0.4\textwidth]{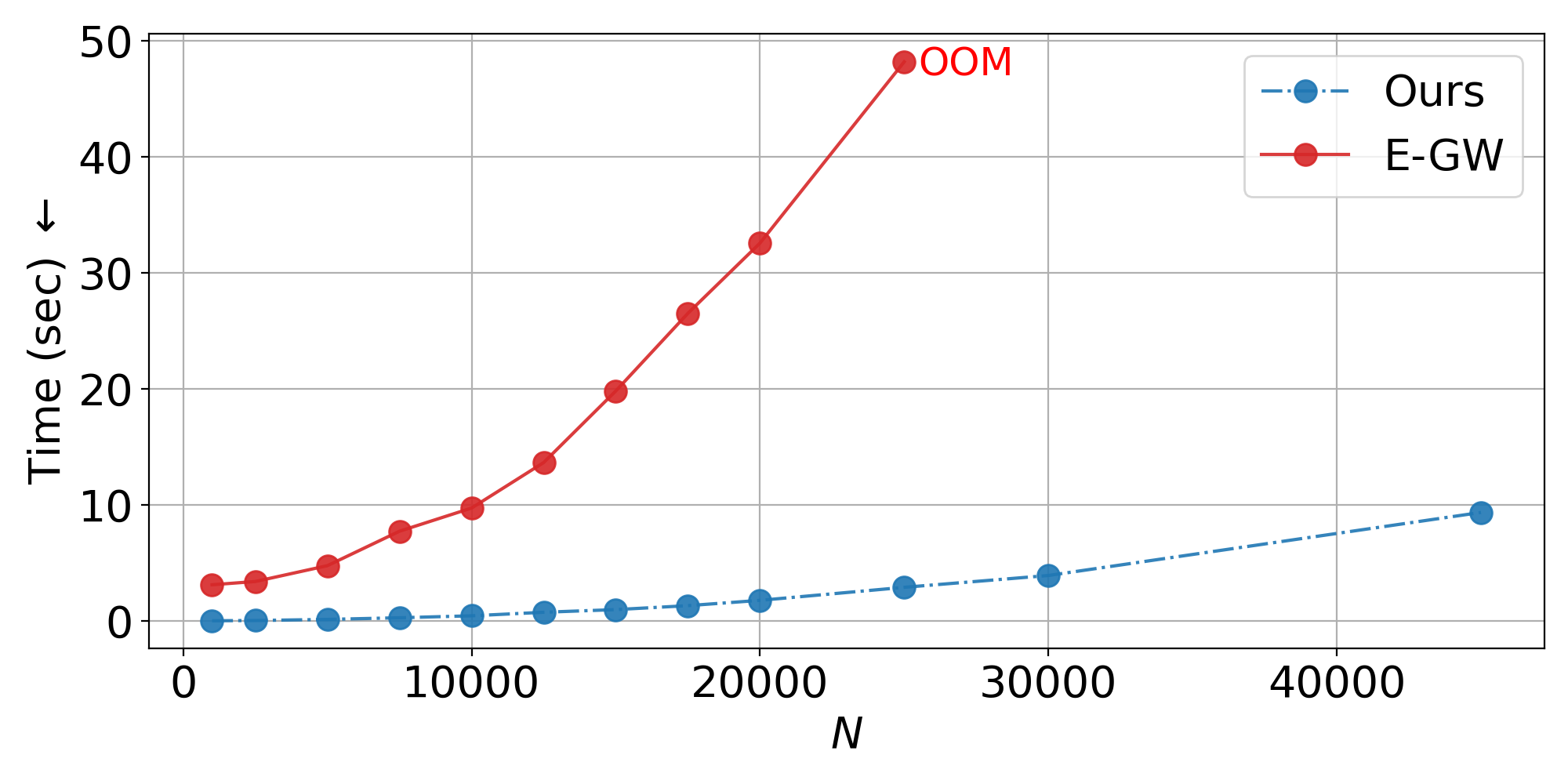}
    \includegraphics[width=0.4\textwidth]{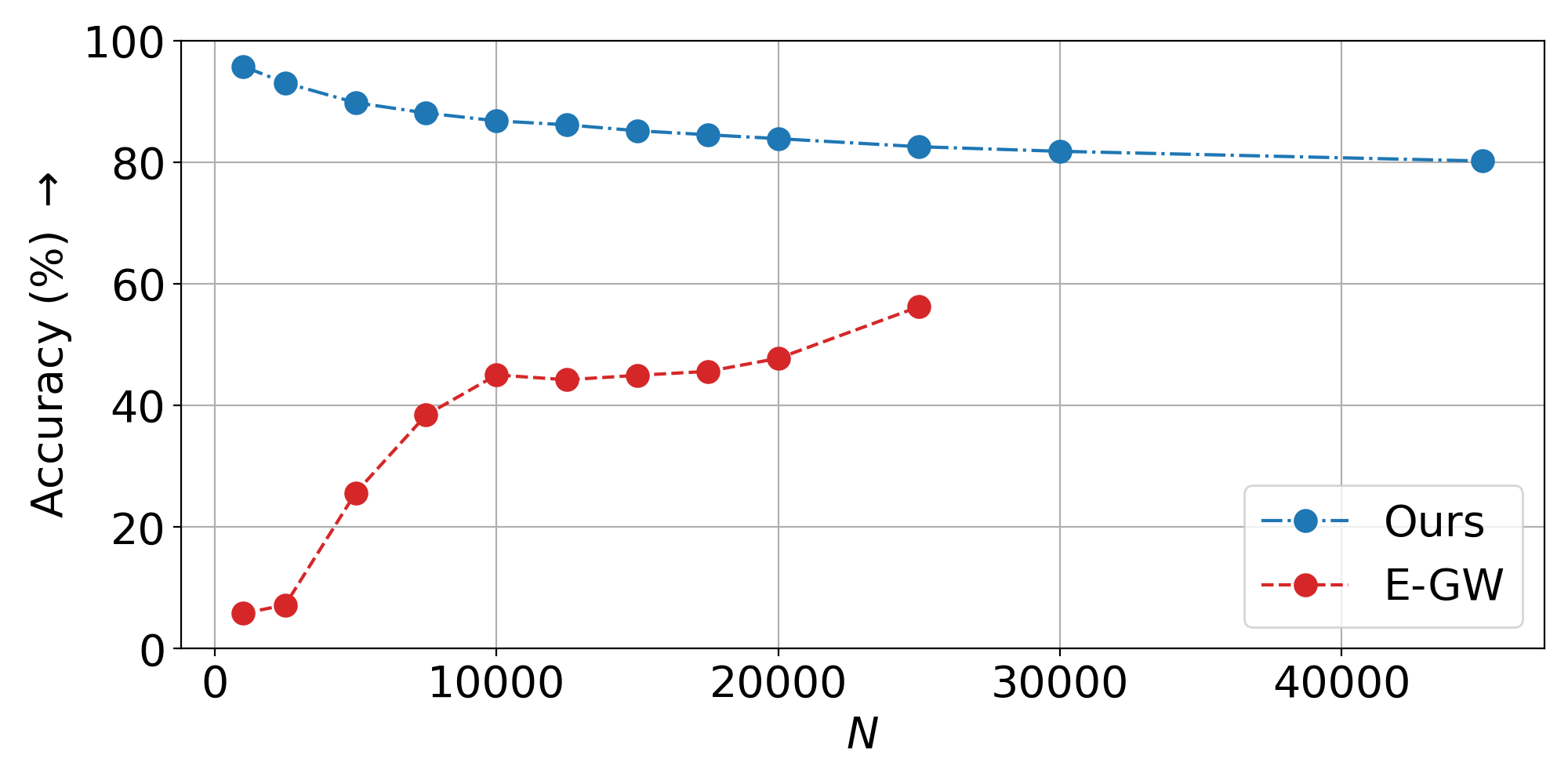}
    \caption{\textbf{The proposed solver generalizes to unseen samples and scales to large-sample sizes post-training.} In both top and bottom experiments, $\mathcal{X}$ and $\mathcal{Y}$ are ViT embeddings. The entropic GW solver can only operate in the transductive regime and runs out of memory for $N>25000$.}
    \label{fig:scale_inductive}
\end{figure}

\begin{figure}
    \centering
    \includegraphics[width=0.25\textwidth]{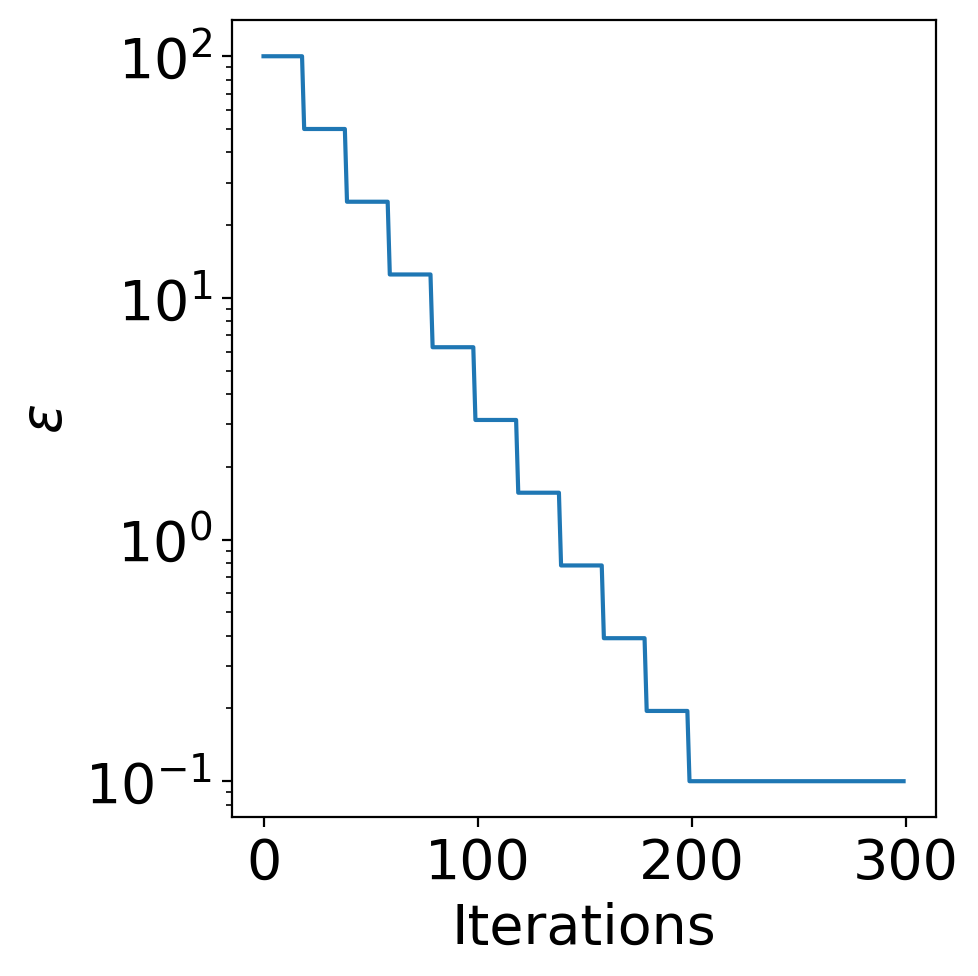}
    \includegraphics[width=0.25\textwidth]{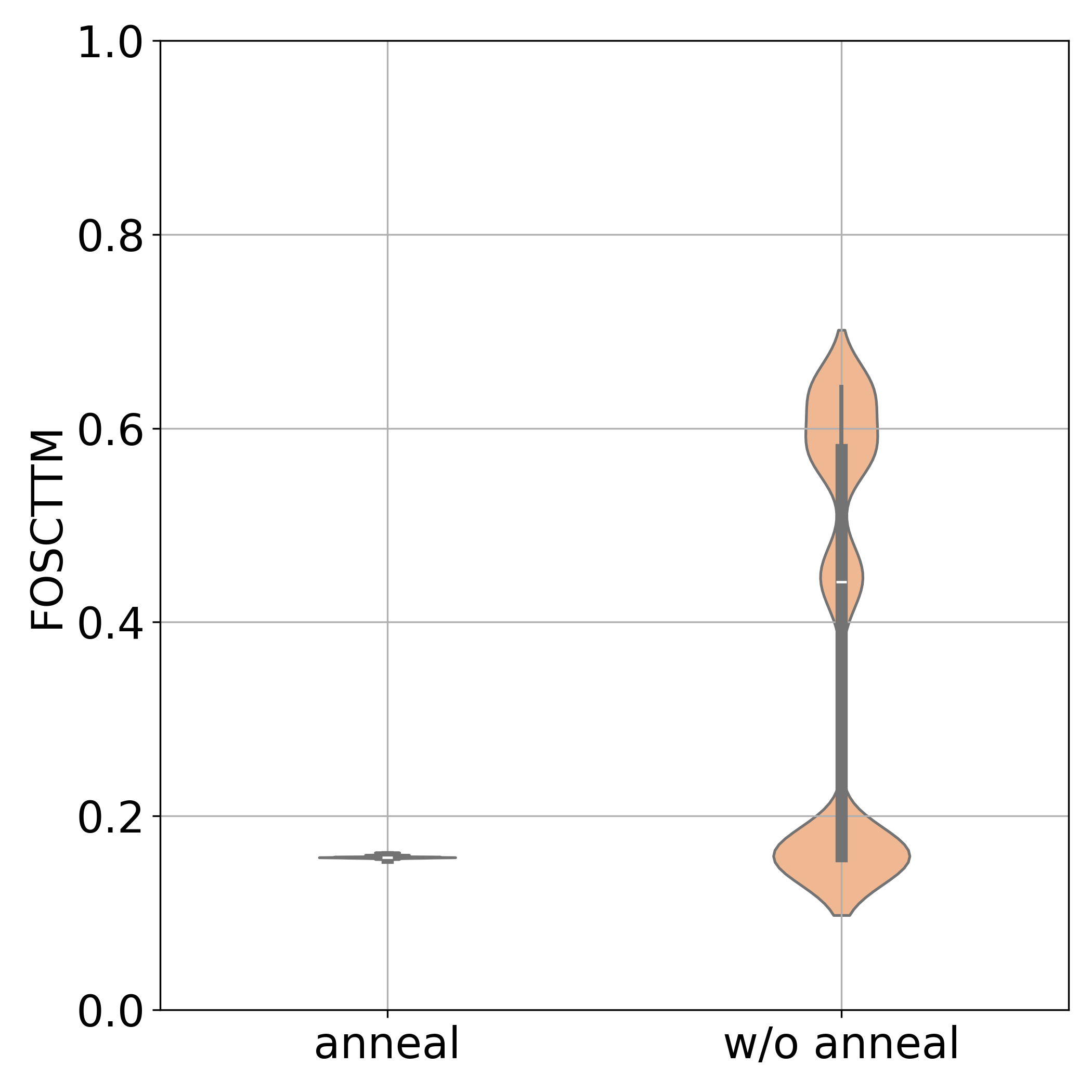}  \includegraphics[width=0.25\textwidth]{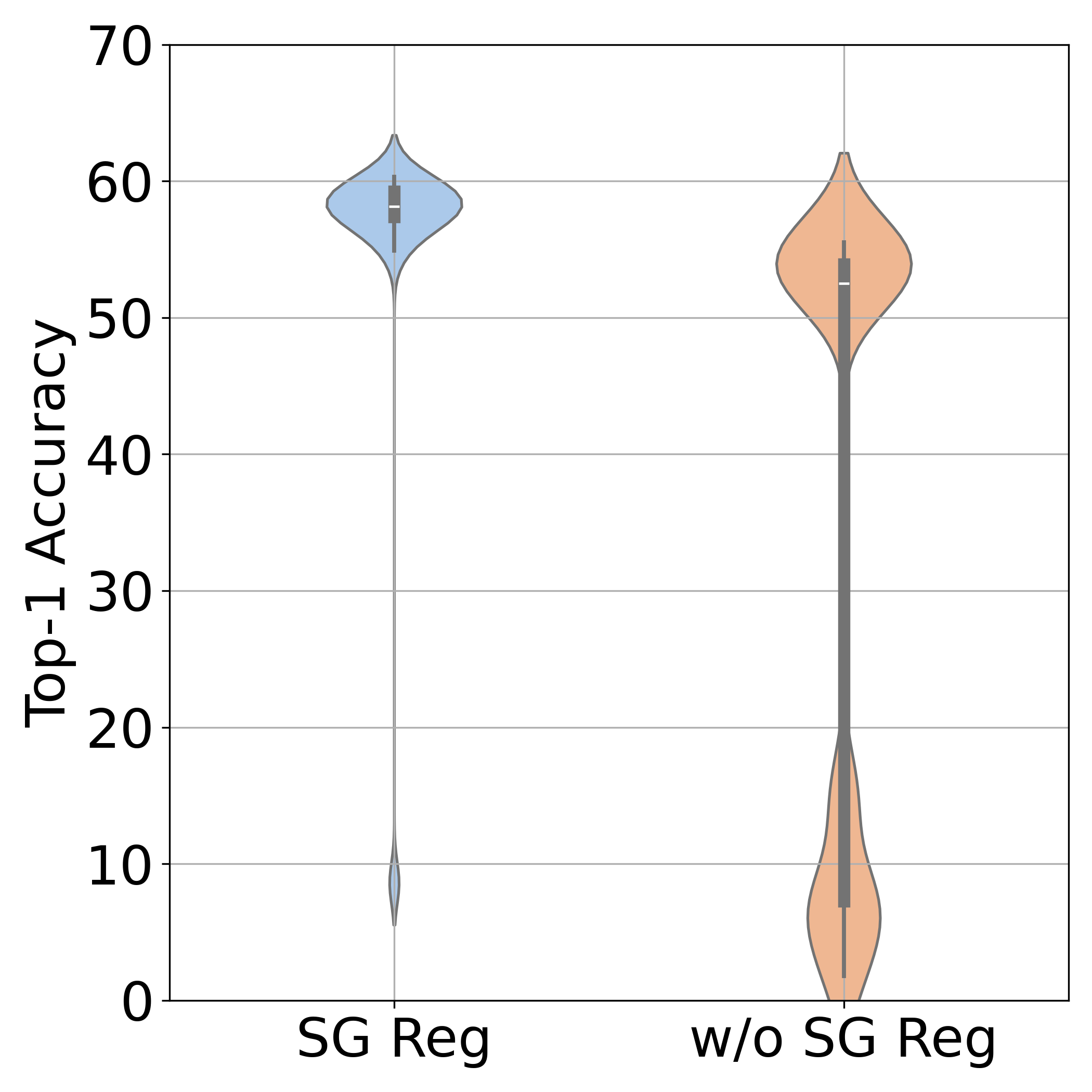}
    \caption{\textbf{Simulated annealing of $\epsilon$ and spectral geometric regularization are effective in stabilizing the solver and improving the accuracy of the assignment.} Left: simulated annealing schedule used. Middle: distribution of the alignment error (measured as FOSCTTM) over $20$ runs with and without $\epsilon$-annealing. Right: distribution of the alignment error with and without the spectral geometric regularization of the transport cost.}
    \label{fig:sim_annealing}
\end{figure}

\section{Extensions}\label{sec:extensions}
While the objective function in Eq. \ref{eq:final_objective_ours} is easy to \textit{evaluate}, the resulting optimization problem is still an NP-hard QAP. In practice, it is challenging to reach good local minima consistently without imposing further inductive biases. This is also true for the entropic GW solver~\cite{peyre2019computational} -- while sequential linearization and projection via Sinkhorn algorithm works reasonably in practice, there exist no guarantees on its global convergence. In fact, there exist many scenarios where it fails to recover a meaningful local minimum. Here we introduce several regularization techniques in the problem in Eq. \ref{eq:final_objective_ours} to remediate poor convergence: (i) \textit{simulated annealing} of the entropic regularization strength $\epsilon$, and (ii) \textit{spectral-geometric regularization of the OT cost}. We also propose a new objective that matches distance ranks instead of distances themselves that can be employed as an alternative Eq. \ref{eq:final_objective_ours}.

\paragraph{Simulated annealing of $\epsilon$.} While evaluating our solver on the scSNARE-seq data~\cite{chen2019high}, where the goal is to align transcriptomic readouts against those of chromatin accessibility and the ground-truth is available thanks to a co-assaying technique developed by \cite{chen2019high}, we observed that our solver, while it is accurate on average, it is sensitive to the initialization of the neural networks $f$ and $g$. As a result of symmetries in the metric spaces of these data, we observed that the assignment sometimes consistently mismapped the cell line of GM12878 to H1, and vice-versa.  
The right panel of Fig. \ref{fig:sim_annealing} depicts the distribution of alignment errors (lower is better) obtained by solving Eq. \ref{eq:final_objective_ours} with multiple random initializations of the embedding parameters $\theta$ and $\phi$. In the right column, the largest mode corresponds indeed to accurate assignment, whereas the two other modes with larger errors represent the aforementioned symmetry-induced cell-line mismappings.

\looseness=-1We mitigate this problem by performing \textit{simulated annealing} on $\epsilon$ of the entropic OT problem within the Sinkhorn layer. We propose a schedule for $\epsilon$ that starts high and is gradually decayed (see Fig. \ref{fig:sim_annealing}, left). Our rationale is that this results in a coarse-to-fine refinement of the learned cost (implicitly parametrized via $f$ and $g$) during training, and it is similar in spirit to the idea of a multi-scale version of kernel matching in shape correspondence problems~\cite{vestner2017efficient, melzi2019zoomout,holzschuh2020simulated}. When $\epsilon$ is high, the entropic regularization is strong, and the resulting assignment is ``softer''. By scheduling $\epsilon$ from a large value to a small one, we demand that the learned cost matrix, and as a consequence, the resulting assignment, gets refined during training. In practice, we observe that the proposed $\epsilon$-scheduling works remarkably well; it practically reduces the variance across seeds to zero and is effective in breaking symmetries in the metric spaces that lead to bad local minima and making the solver more reliable (Fig.~\ref{fig:sim_annealing}, middle). 

\looseness=-1\paragraph{Spectral representation on graphs.} Before introducing our proposed spectral-geometric regularization, we provide a brief background on graphs. A familiar reader may skip to the following paragraph.
Let $\mathcal{G} = (V, E, \mathbf{\Omega})$ be a weighted graph with the vertex set $V$, edge set $E$, and weighted adjacency matrix $\mathbf{\Omega}$. The \textit{combinatorial graph Laplacian} is defined as $\mathbf{L} = \mathbf{D} - \mathbf{\Omega}$, where $\mathbf{D} = \textrm{diag}(\mathbf{\Omega}\vec{1})$ is \textit{the degree matrix}. Given a scalar-valued function $\mathbf{z} \in \mathbb{R}^{|V|}$ on the graph $\mathcal{G}$, the \textit{Dirichlet energy} is defined to be $\mathbf{z}^\T\mathbf{L}\mathbf{z}$, and it measures the \textit{smoothness} of $\mathbf{z}$ on $\mathcal{G}$ \cite{spielman2012spectral}. 
Given two graphs $\mathcal{G}_1 = (V_1, E_1, \mathbf{\Omega}_1)$ and $\mathcal{G}_2 = (V_2, E_2, \mathbf{\Omega}_2)$, the Cartesian product of $\mathcal{G}_1$ and $\mathcal{G}_2$, denoted by $\mathcal{G}_1\,\Box\,\mathcal{G}_2$, is defined as a graph with the vertex set $|V_1| \times |V_2|$, on which two nodes $(u, v), (u', v')$ are adjacent if either $u=u'$ and $(v, v) \in E_2$ or $v = v'$ and $(u, u) \in E_1$.
The Laplacian of $\mathcal{G}_1\,\Box\,\mathcal{G}_2$ is defined as a tensor sum of the Laplacians $\mathbf{L}_1$ and $\mathbf{L}_2$, i.e., $\mathbf{L}_{\mathcal{G}_1 \Box \mathcal{G}_2} = \mathbf{L}_1 \oplus \mathbf{L}_2 = \mathbf{L}_1 \otimes \mathbf{I} + \mathbf{I} \otimes \mathbf{L}_2$. 
Denote the spectral decompositions of the Laplacians by $\mathbf{L}_1 = \vec{\Phi} \vec{\Lambda}_1 \vec{\Phi}^\T$ and $\mathbf{L}_2 = \vec{\Psi} \Vec{\Lambda}_2 \Vec{\Psi}\T$. A signal $\mathbf{Z}$ on the product graph $\mathcal{G}_1\,\Box\,\mathcal{G}_2$ can be represented using the bases of the individual Laplacians as $\mathbf{Z} = \Vec{\Phi}\T \mathbf{F} \Vec{\Psi}$, with the coefficients $\mathbf{F}$.

\paragraph{Spectral-geometric regularization of the OT cost.} We propose a spectral-geometric regularization on the learned OT cost that demands ``similar'' items in $\mathcal{X}$ to incur ``similar'' cost with respect to all items in $\mathcal{Y}$, and vice-versa. To formally represent this notion, let $\mathcal{G}_\mathcal{X} = (\mathcal{X}, \mathcal{E}_\mathcal{X}, \mathbf{\Omega}_\mathcal{X})$ and $\mathcal{G}_\mathcal{Y} = (\mathcal{Y}, \mathcal{E}_\mathcal{Y}, \mathbf{\Omega}_\mathcal{Y})$, be two graphs \textit{inferred} on $\mathcal{X}$ and $\mathcal{Y}$, respectively, and let $\mathbf{L}_\mathcal{X}$ and $\mathbf{L}_\mathcal{Y}$ be their corresponding graph Laplacians. We interpret the learned OT cost $\mathbf{C}=c(f_\theta(\mathbf{X}), g_\phi(\mathbf{Y}))$ from Eq. \ref{eq:final_objective_ours} as a signal on the product graph $\mathcal{G}_\mathcal{X}\,\Box\,\mathcal{G}_\mathcal{Y}$, and demand that $\mathbf{C}$ is \textit{smooth} on $\mathcal{G}_\mathcal{X} \, \Box \, \mathcal{G}_\mathcal{Y}$. The latter smoothness can be expressed as the Dirichlet energy of $\mathbf{C}$ measured on $\mathcal{G}_\mathcal{X} \, \Box \, \mathcal{G}_\mathcal{Y}$,
\begin{equation}
    \label{eq:dir_energy}
    \mathcal{E}_{\text{sm}} = \mathrm{trace}\left( \mathbf{C}^\T\left(
    \mathbf{L}_\mathcal{X} \otimes \mathbf{I} + \mathbf{I} \otimes \mathbf{L}_\mathcal{Y} 
    \right)\mathbf{C} \right) = 
    \mathrm{trace}\left( \mathbf{C}^\T \mathbf{L}_{\mathcal{X}} \mathbf{C} + 
    \mathbf{C} \mathbf{L}_\mathcal{Y} \mathbf{C}^\T \right),
\end{equation}
and added to Eq. \ref{eq:final_objective_ours} as an additional regularization.  Figure \ref{fig:sim_annealing} (right) demonstrates the effectiveness of the proposed spectral regularization on the task of aligning embeddings from neural latent spaces.

From the spectral perspective, interpreting the OT cost $\mathbf{C}$ as a signal on $\mathcal{G}_\mathcal{X}\,\Box\,\mathcal{G}_\mathcal{Y}$, learning $\mathbf{C}$ given the pointwise features from domains $\mathcal{X}$ and $\mathcal{Y}$ is equivalent to directly learning the functional map of $\mathbf{C}$, this makes our work intimately related to the works of that learn functional maps~\cite{litany2017deep, halimi2019unsupervised, vestner2017efficient, boyarski2022spectral, kalofolias2014matrix} from shape correspondence and geometric matrix completion literature.

\paragraph{Matching ranks instead of distances.} The choice of the comparison criterion for the pairwise distances crucially influences the usability of the GW problem for real applications. Consider, for example, two point clouds that differ only by a scale factor; since distances are not scale-invariant, solving Eq. \ref{eq:final_objective_ours} to match distances would produce meaningless results. As a remedy, we propose to match the \textit{ranks} of the pairwise distances instead of their absolute values. Ranks preserve the order and are insensitive to scale or, more generally, monotone transformations. This departs from the standard framework of GH and GW, which align metric spaces, and generalizes it to a more general problem of performing unpaired alignment by matching non-metric quantities. In order to be able to differentiate the objective with respect to ranks, which is an inherently non-differentiable function, we use the differentiable soft ranking operators introduced by \citet{blondel2020fast}.
We optimize the following modified objective:
\begin{equation}
\begin{aligned}
    \label{eq:final_objective_ours_rank}
    \Vec{\Pi}^* =& \argmin_{\theta, \phi}  \, 
    \left\| 
    \mathcal{R}_\delta\left(\mathbf{D}_\mathcal{X}\right) - \mathcal{R}_\delta\left(\Vec{\Pi}(\theta, \phi) \mathbf{D}_\mathcal{Y} \Vec{\Pi}^\T(\theta, \phi) \right) 
    \right\|_{\mathrm{F}}^2 \\
    & \,\, \text{  s.t.  } \Vec{\Pi}(\theta, \phi) = \argmin_{\Vec{\Pi} \in U(\mu, \nu)} \langle \mathbf{\Pi}, {c}(f_\theta(\rvec{X}), g_\phi(\rvec{Y}))\rangle,
\end{aligned}
\end{equation}
where $\mathcal{R}_\delta$ is a soft-ranking operator applied separately to each row of the matrix, and $\delta$ controls the level of ``softness'' of the rank. Because ranking is a nonlinear operation, this results in a problem that is no longer quadratic in $\mathbf{\Pi}$, it is unclear how standard GW solvers can be adapted to such settings, and also highlights the benefit of having a gradient-descent--based solver.
Applying ranking to other groups of distances effectively results in a different GW-like distance. We defer the systematic exploration of this new family of distances to future work.

\textbf{Further extensions.} Although we do not explore it within this work, it is easy to see that (i) the proposed framework can be extended to a fused GW~\cite{fusedgw} setting by adding a linear objective to Eqs. \ref{eq:final_objective_ours} and \ref{eq:final_objective_ours_rank}; (ii) the rank of the OT cost can be controlled by modifying the dimension of the embeddings' output by $f$ and $g$; and (iii) when partial supervision is available on the assignment (``semi-supervised'' alignment), it can be incorporated into the loss as a {data term}.

\section{Experiments}
We split this experiment section into three parts. 
Firstly, we demonstrate that our solver works in the inductive setting and that is much more scalable to large sample sizes in this setting. Secondly, we showcase experiments that demonstrate the effects of (i) \textit{simulated annealing of $epsilon$}, (ii) \textit{spectral geometric regularization}, and (iii) the \textit{ranking-based} formulation. Thirdly, we demonstrate that the proposed solver, in the transductive setting, outperforms the entropic GW solver on two single-cell multiomics benchmarks. We use both real and synthetic data wherever appropriate.

\textbf{Inductivity and scale. \ } In order to evaluate the inductivity of the method and to benchmark it against the entropic GW solver, we consider two experiments (i) when  $\mathcal{X}$ and $\mathcal{Y}$ are \textit{isometric}, and (ii) when $\mathcal{X}$ and $\mathcal{Y}$ are not exactly isometric. 
For the \textbf{first experiment}, we consider $\mathcal{X}$ to be CIFAR100 encodings obtained from a vision transformer~\cite{dosovitskiy2020image}. We apply an orthogonal transformation to each element of $\mathcal{X}$ to generate $\mathcal{Y}$. We parametrize our encoders $f$ and $g$ to be $3$-layer multi-layer perceptrons (MLPs), and optimize the Eq. \ref{eq:final_objective_ours} with respect to their parameters on $200$ unaligned samples for $500$ iterations ($12$ seconds). Then, we evaluate our method in an \textit{inductive setting} with an increasing number of unaligned samples available at inference up to $N=45000$. We benchmark it against the GPU-accelerated entropic GW solver available from \texttt{ott-jax}~\cite{cuturi2022optimal}. The results are presented in the top panels of Figure \ref{fig:scale_inductive}. We measure accuracy as whether the predicted correspondence is correct \textit{in terms of the class label}. We observe that both solvers recover the orthogonal transformation perfectly. Further, we can observe that an inductive solver, because it solves only a $\epsilon$-OT problem at inference, is much faster and more memory efficient. Employing an entropic GW solver, on the other hand, goes out of memory for $N>25000$. Note that the times we reported \textit{do not include} the time required to compute a geodesic distance matrix for both $\mathcal{X}$ and $\mathcal{Y}$, which is significantly time-consuming at large sample sizes (>10 mins for $N=20000$). In contrast, using our solver would not require computing $\mathbf{D}_\mathcal{X}$ and $\mathbf{D}_\mathcal{Y}$ at inference. For the \textbf{second experiment}, we use the data from \cite{maiorca2024latent} and choose $\mathcal{X}$ to be ViT embeddings as in the previous experiment, while $\mathcal{Y}$ is set to be ViT embeddings generated from \textit{rescaled} images. We train the $f$ and $g$ for $1000$ iterations ($\sim 2$ minutes), using $1000$ unpaired samples during the training time. The results are presented in the bottom two panels of Fig.\ref{fig:scale_inductive}. The results suggest, again, that our solver both generalizes well and scales gracefully with sample sizes, whereas the entropic-GW solver produced inferior results in this setting. These experiments corroborate our claim that our solver both attains high-quality solutions and scales well in the inductive regime.

\textbf{Spectral geometric regularization. \ } For this experiment, we consider the above setting where $\mathcal{X}$ and $\mathcal{Y}$ are two unaligned sets of embeddings obtained from a pre-trained vision transformer~\cite{dosovitskiy2020image}. We set $f$ and $g$ to be 3-layer MLPs solve Eq. \ref{eq:final_objective_ours} with and without $\mathcal{E}_{\text{sm}}$ regularization (Eq. \ref{eq:dir_energy}). 
We solve this problem on $20$ unaligned datasets drawn from $\mathcal{X}$ and $\mathcal{Y}$, each of size $N=1000$. Figure \ref{fig:sim_annealing} (right panel) presents the accuracy of the assignment by measuring if the predicted corresponding point belongs to the same class as the groundtruth correspondence. Notice that geometric regularization improves the accuracy of the assignment (+20\% in terms of mean accuracy over trials). Moreover, it also reduces the variance thereby inducing meaningful inductive bias into the solver.

\textbf{Simulated annealing of $\epsilon$. \ } As discussed in Section \ref{sec:extensions}, we consider the scSNARE-seq data~\cite{snareseq}, which is a co-assay of transcriptome and chromatin accessibility measurements performed on $N=1047$ cells. We run our experiment with and without the proposed simulated annealing of $\epsilon$ for 20 random initializations of $f$ and $g$, the results are presented in Figure \ref{fig:sim_annealing}. We observe that using this seed stabilizes the training process significantly. We used this as a default choice across all real data experiments. 

\begin{figure}
    \centering
    \includegraphics[width=0.8\textwidth]{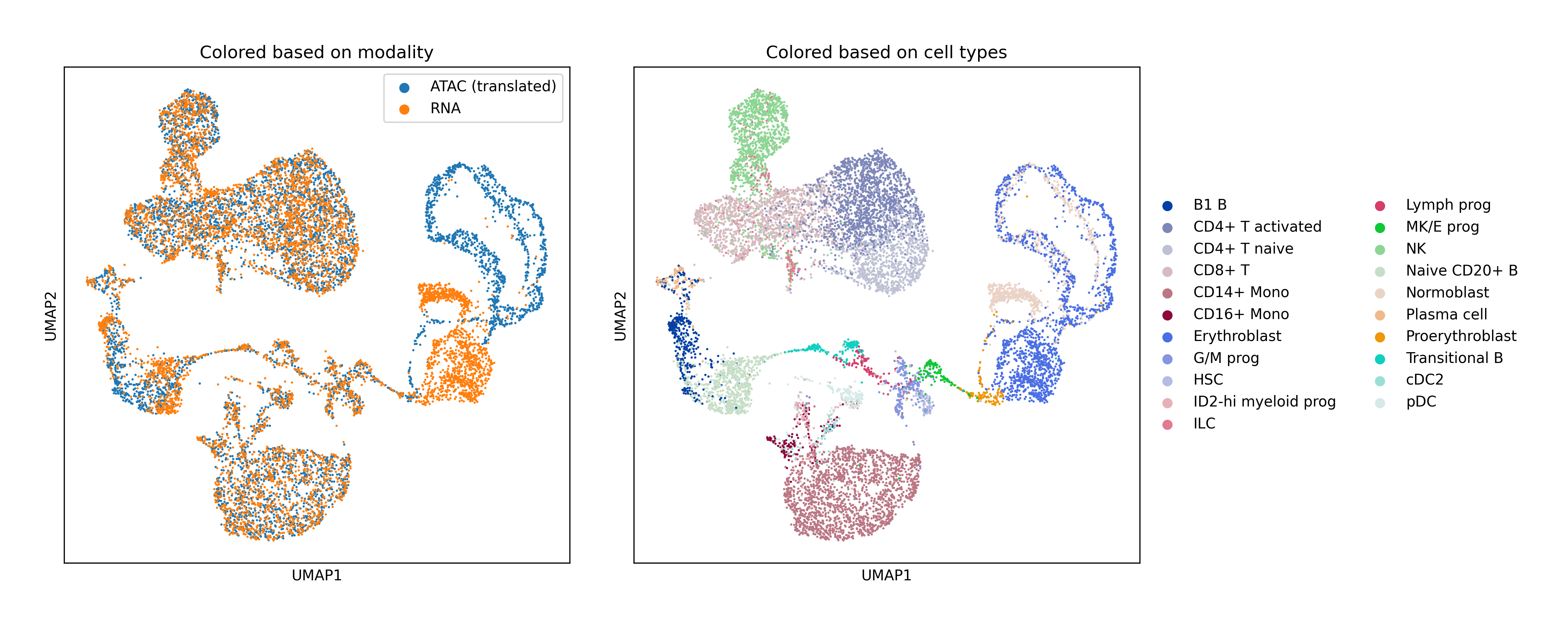}\\
    \includegraphics[width=0.4\textwidth]{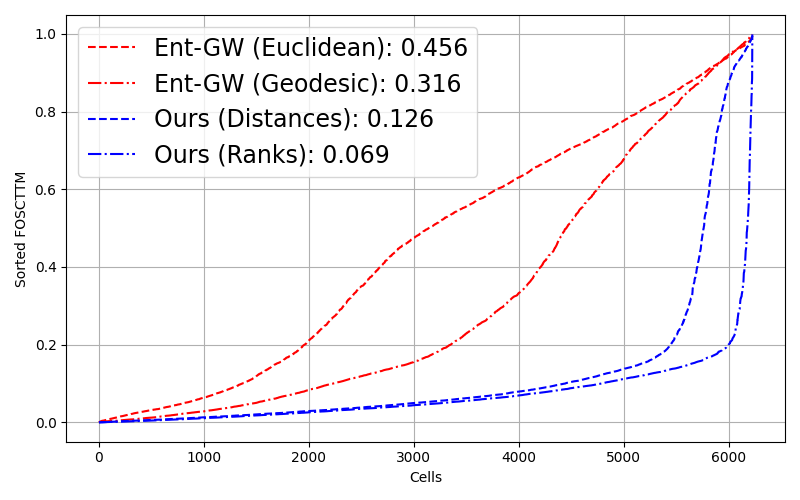}
    \includegraphics[width=0.4\textwidth]{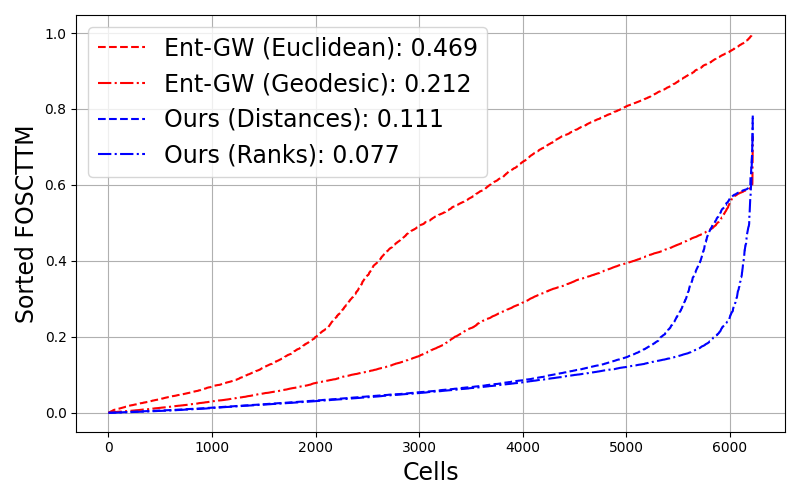}
    \caption{\textbf{Qualitative and quantitative results on the human bone marrow single-cell dataset.} Top plots depict the UMAP of the translated cells colored by domain (left) and by the cell type (right).Bottom plots report the FOCSTTM metrics for $\mathcal{Y}$ projected onto $\mathcal{X}$ (left) and $\mathcal{X}$ projected onto $\mathcal{Y}$ (right). }
    \label{fig:moscot}
\end{figure}

\textbf{Ranking-based GW. \ } Figure \ref{fig:swiss_roll} (right panel) depicts the assignment produced by the ranking-based GW solver in an inductive setting. 
We observe that ranking-based GW outperforms the distance-based counterpart in the setting of single-cell multiomic alignment. Consequently, the results that we present in the sequel (Figures \ref{fig:moscot} and \ref{fig:scot}) use ranking-based loss and they outperform both the entropic GW solver and the distance-based variant of our solver.

\textbf{Single-cell multiomic alignment. \ } We consider two real-world datasets: (i)\textbf{ scSNARE-seq} data which contains gene expression (RNA) and chromatin accessibility (ATAC) profiles form 1047 individual cells from four cell lines: H1, BJ, K562, and GM12878, with known groundtruth thanks to a co-assaying technique developed by \cite{snareseq}. We obtained the processed data of RNA and ATAC features from the \citet{demetci2022scot}, whose method uses entropic GW to align these two modalities and serves as the baseline we evaluate against.
(ii) \textbf{human bone marrow} single-cell dataset that contains \textit{paired} measurements of single-cell RNA-seq and ATAC-seq measurements released by \citet{moscotdata}. We obtained the processed data from \texttt{moscot}~\cite{klein2023moscot}. In the RNA space, we used PCA embedding of 50 dimensions, and in the ATAC space, we used an embedding given by LSI (latent semantic indexing) embedding, followed by $L_2$ normalization.

In the scSNARE-seq experiment, we used the entropic GW solver with the same hyperparameters used by \cite{demetci2022scot} as the baseline. It was shown by \cite{demetci2022scot} to outperform the other baselines for unpaired alignment on this data. 
In the bone marrow single-cell experiment, we compared to the entropic GW solver with Euclidean metric and the geodesic distance metric. To establish a fair baseline, following the methodology of \citet{demetci2022scot}, for both baselines, we perform a grid search on the $\epsilon$ used in Sinkhorn iterations of the solver, and $k$ corresponding to the $k$-NN graph constructed for geodesic computation (for the latter setting), and pick the hyperparameters with the least GW loss. In the case of both bone marrow data and the scSNARE-seq data, we observe that our ranking-based solver produces the best FOCSTTM score (see Appendix). In the case of bone marrow data, especially, our solver produces a significant margin over the entropic GW solvers. The results of scSNARE-seq alignment are presented in Figure \ref{fig:scot} in the Appendix. In scSNARE-seq, the margin of our improvement is lower, this could be attributed to limited diversity in cell-lines and small sample-size in scSNARE-seq compared to the bone-marrow data.

\section{Conclusion} 
In this paper, we presented a new scalable approach to the Gromov-Wasserstein problem. The GW loss is pair-wise and thus is hard to minimize directly yet simple to evaluate. On the other hand, the OT loss is point-wise and is thus simple to minimize efficiently. We showed practical approaches to learning data embeddings such that the solution of the corresponding OT problem minimizes the GW loss. Unlike existing GW solvers that optimize the assignment matrix or the corresponding dual variables directly, our optimization variables are the parameters of the embedding functions. In addition to better scalability in the transductive regime, the proposed approach is also inductive, as the computed embeddings can be applied to new data previously unseen in training. We further proposed regularization techniques demonstrating consistently better convergence. We emphasize that GW is an NP-hard problem, and no existing polynomial-time algorithms (including ours) are guaranteed to find its global minimum. However, we showed in many synthetic and real data experiments that the proposed solver is significantly more accurate and scalable. 

We also introduced a new distance between metric-measure spaces in which distance ranks are matched instead of the distances themselves, which is more appropriate for metric structures coming from distinct modalities that do not necessarily agree quantitatively. 
Being oblivious to any monotone transformation of the metric structure, this new distance can be applied to general non-metric dissimilarities in the spirit of non-metric multidimensional scaling (MDS) \cite{mdscox}. We defer to future studies the exploration of its geometric and topological properties.

\looseness=-1\paragraph{Limitations.} Our current approach focuses on the discrete GW problem in which the correspondence is found explicitly. Future work should study the continuous setting, with the correspondence represented, e.g., in the form of a functional map \cite{ovsjanikov2012functional} -- an operator mapping functions on to $\mathcal{X}$ to functions on $\mathcal{Y}$ which can be represented efficiently using truncated bases of the product graph constructed on $\mathcal{X} \times \mathcal{Y}$. Another limitation is the use of full batches for the minimization of the GW loss, which restricts scalability in the transductive regime. Future studies should consider extending the proposed approach to the mini-batch setting, in the spirit of mini-batch optimal flow-matching \cite{tong2023improving}, \cite{klein2023generative}.

\bibliography{neurips_2024}

\newpage
\appendix

\section{Appendix}

\paragraph{Sinkhorn algorithm.} The Sinkhorn algorithm allows efficient solution of the entropy-regularized linear OT problem of the form
$$
\min_{\mathbf{\Pi} \in U(\mu, \nu)} \langle \mathbf{C}, \mathbf{\Pi} \rangle + \epsilon \langle \mathbf{\Pi}, \log \mathbf{\Pi} \rangle.
$$
Defining the kernel matrix $\mathbf{K} = e^{-\mathbf{C} / \epsilon}$ and initializing $\mathbf{u}_1 = \mathbf{v}_1 = \mathbf{1}$, the algorithm proceeds with iterating
$$
\mathbf{u}_{k+1} = \frac{\mu}{ \mathbf{K} \mathbf{v}_k };  \hspace{1cm}
\mathbf{v}_{k+1} = \frac{\nu}{ \mathbf{K}\T \mathbf{u}_{k+1} },
$$
from which the assignment matrix $\mathbf{\Pi}_{k+1} = \mathrm{diag}(\mathbf{u}_{k+1}) \, \mathbf{K} \, \mathrm{diag}(\mathbf{v}_{k+1})$. Here $\mathrm{diag}(\mathbf{u})$ denotes a diagonal matrix with the entries of the vector $\mathbf{u}$ on the diagonal, and exponentiation and division are performed element-wise. The iterations are usually stopped when the change $\|\mathbf{\Pi}_{k+1} - \mathbf{\Pi}_{k}\|$ falls below a pre-defined threshold.

\paragraph{Entropic GW solver.} 
The entropic GW solver aims at solving the entropy-regularized GW problem
$$
\min_{\mathbf{\Pi} \in U(\mu, \nu)} \|\mathbf{D}_\mathcal{X} - \mathbf{\Pi} \mathbf{D}_\mathcal{Y} \mathbf{\Pi}\T\|_\mathrm{F}^2 + \epsilon \langle \mathbf{\Pi}, \log \mathbf{\Pi} \rangle.
$$
Without the entropy term, the problem is a linearly constrained quadratic program, which
Gold and Rangarajan \cite{gradassgn} proposed to solve as a sequence of linear programs. Applied here, this idea leads to a sequence of entropy-regularized linear OT problems of the form
$$
\mathbf{\Pi}_{k+1} = \mathrm{arg}\min_{\mathbf{\Pi} \in U(\mu, \nu)} \langle \mathbf{C}_{k+1}, \mathbf{\Pi} \rangle + \epsilon \langle \mathbf{\Pi}, \log \mathbf{\Pi} \rangle,
$$
with the cost $\mathbf{C}_{k+1} = \mathbf{D}_\mathcal{X}\mathbf{\Pi}_{k} \mathbf{D}_\mathcal{Y}$ defined using the previous iteration. Each such problem is solved using Sinkhorn inner iterations. 

\paragraph{Barycentric projection.} For visualization and comparison purposes, it is often convenient to represent the points from $\mathcal{X}$ and $\mathcal{Y}$ in the same space. Let $\mathbf{X} = (\mathbf{x}_1,\dots,\mathbf{x}_N)$ and $\mathbf{Y} = (\mathbf{y}_1,\dots,\mathbf{y}_N)$ denote the coordinates of the points in $\mathcal{X}$ and $\mathcal{Y}$, respectively. Given the ``soft" assignment $\mathbf{\Pi}$ and using $\mathcal{Y}$ as the representation space, we can represent $\mathbf{X}$ in the form of the weighted sum, $\hat{\mathbf{X}} = \mathbf{Y} \mathbf{\Pi}$, so that the representation of a point $\mathbf{x}_i$ in $\mathbf{Y}$ becomes \cite{alvarez2018gromov}
$$
\hat{\mathbf{x}}_i = \sum_{j} \pi_{ij} \mathbf{y}_j,
$$
We remind that $\mathbf{\Pi}$ is by definition a stochastic matrix, implying that the weights in the above sum are non-negative and sum to $1$.

\paragraph{FOSCTTM score.} The \emph{fraction of samples closer than the true match} (FOSCTTM) measures the alignment quality of two equally-sized sets with known ground-truth correspondence. Let $U = \{\mathbf{u}_i  \}$ and $V = \{\mathbf{v}_i \}$ be two sets of points in a common metric space $\mathcal{Z}$ ordered, without loss of generality, in trivial correspondence order (i.e., every $\mathbf{u}_i$ corresponds to $\mathbf{v}_i$). 
Given a point $\mathbf{u}_i$, we define the fraction of points in $V$ that are closer to it than the true match $\mathbf{v}_i$,
$$
p_i = \frac{1}{N} \, \left|\{ j : d_\mathcal{Z}(\mathbf{u}_i, \mathbf{v}_j) < d_\mathcal{Z}(\mathbf{u}_i, \mathbf{v}_i) \} \right|.
$$
Similarly, we define the fraction of points in $U$ that are closer to $\mathbf{v}_i$ and the true match $\mathbf{u}_i$,
$$
q_i = \frac{1}{N} \, \left|\{ j : d_\mathcal{Z}(\mathbf{v}_i, \mathbf{u}_j) < d_\mathcal{Z}(\mathbf{v}_i, \mathbf{u}_i) \} \right|.
$$
The FOCSTTM score is defined as 
$$
\mathrm{FOCSTTM} = \frac{1}{2N} \sum_{i=1}^N (p_i + q_i).
$$
The score is normalized in the range of $[0,1]$ with perfect alignment having $\mathrm{FOCSTTM} = 0$.

\begin{figure}
    \centering
    \includegraphics[width=0.75\textwidth]{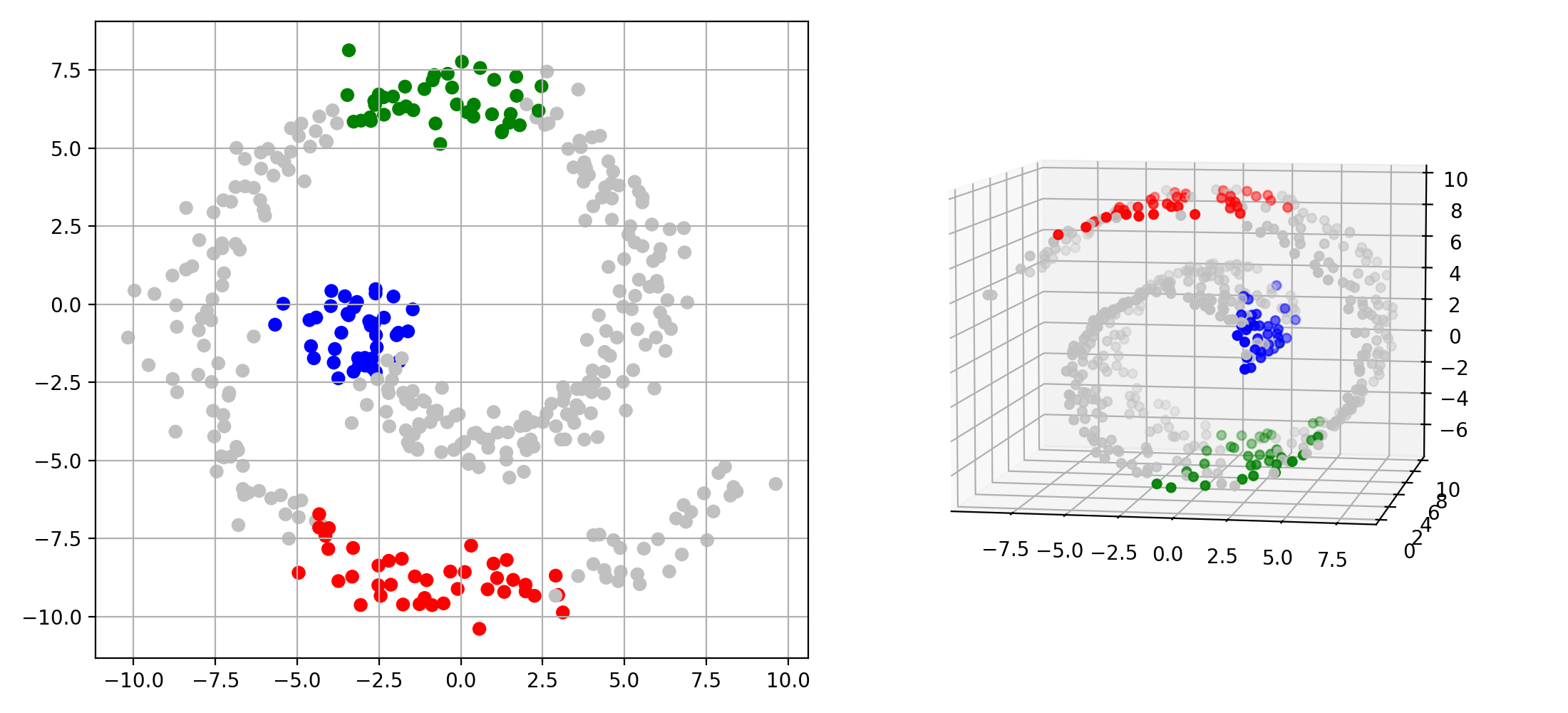}
    \caption{\textbf{Qualitative evaluation of the proposed GW solver in inductive setting. } The plot depicts the assignment produced by our distance-based GW solver (Eq. \ref{eq:final_objective_ours}) on a new set of samples.}
    \label{fig:swiss_roll}
\end{figure}

\begin{figure}
    \centering
    \includegraphics[width=0.3\textwidth]{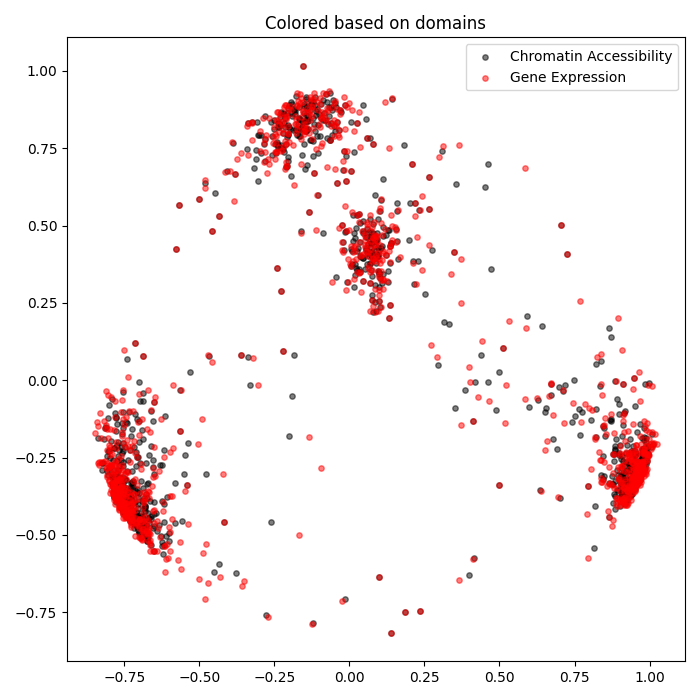}
    \includegraphics[width=0.3\textwidth]{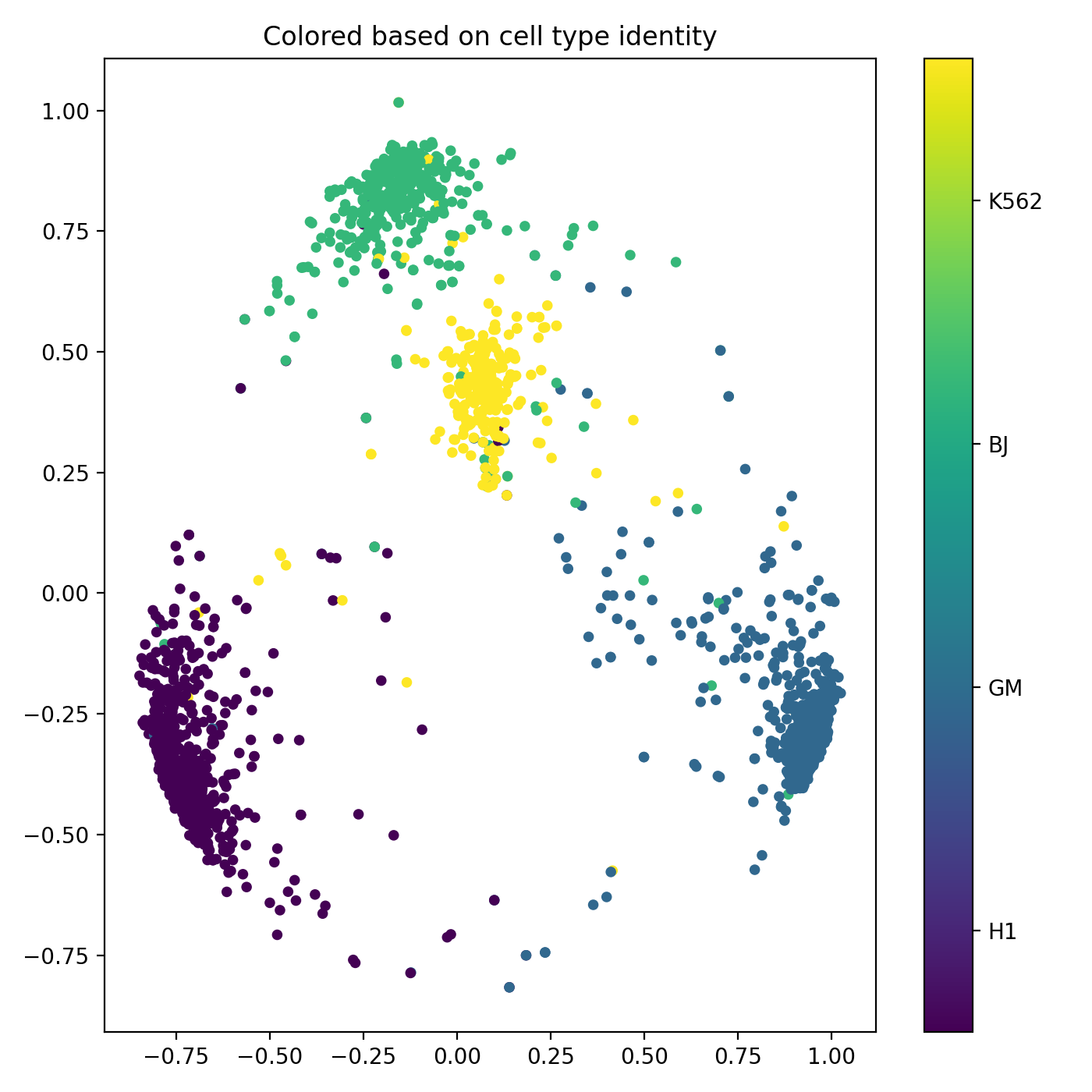}
    \includegraphics[width=0.3\textwidth]{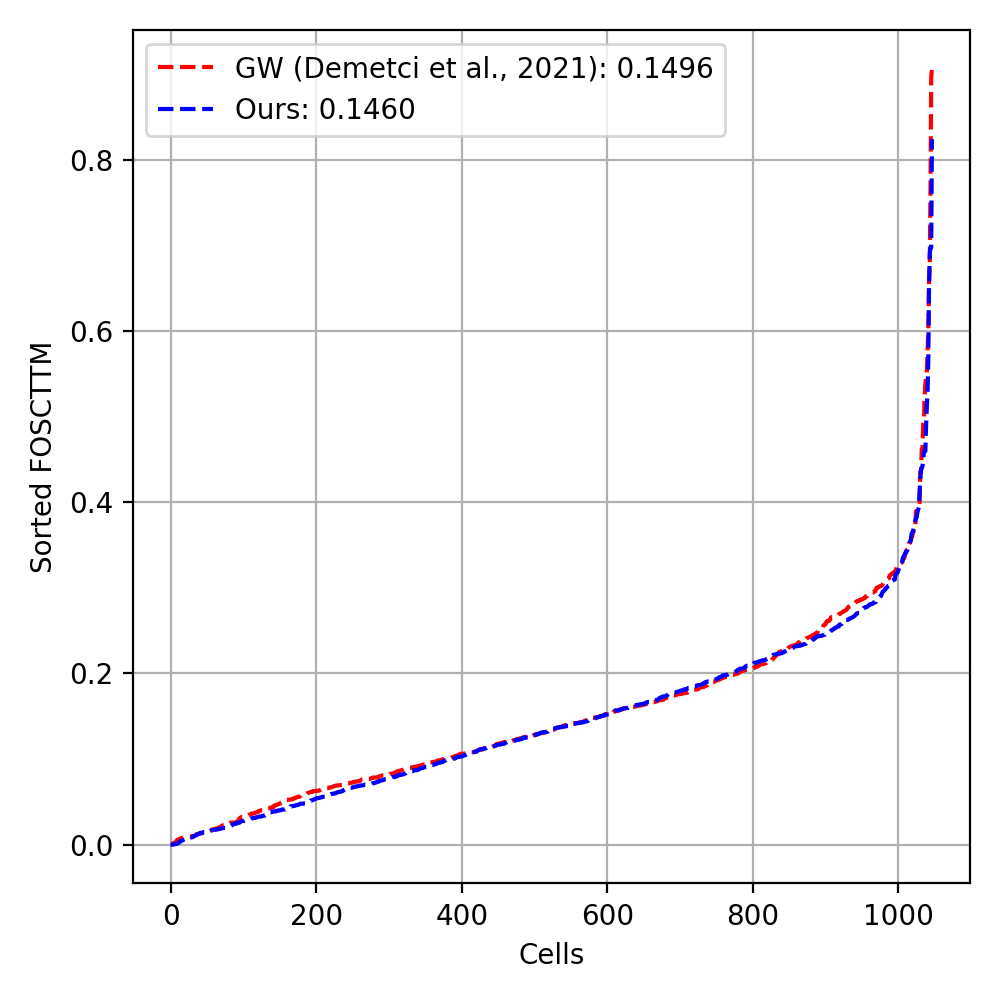}
    \caption{\textbf{Qualitative and quantitative results on the scSNARE-seq dataset.} Left and middle: Aligned samples from ATAC and RNAl, colored by the domains (ATAC: black, RNA: red) and cell types, respectively. Right: the sorted FOCSTTM plot, a quantitative metric measuring the quality of the assignment.}
    \label{fig:scot}
\end{figure}


\end{document}